%% file: main.tex
\documentclass[
hf,
]{ceurart}

\usepackage{listings}
\usepackage[most]{tcolorbox}
\usepackage{fancyvrb}
\usepackage{xcolor}

\definecolor{rolecol}    {HTML}{1F4E79}  
\definecolor{actcol}     {HTML}{B7950B}  
\definecolor{rulecol}    {HTML}{922B21}  
\definecolor{statecol}   {HTML}{1E8449}  
\definecolor{exmpcol}    {HTML}{6C3483}  
\definecolor{outcol}     {HTML}{117A65}  

\definecolor{enrichbg}   {HTML}{FDEBD0}  
\definecolor{statebg}    {HTML}{D4E6F1}  
\definecolor{ratbg}      {HTML}{F5B7B1}  

\newtcolorbox{promptsec}[3]{%
  enhanced, breakable, boxrule=0pt, arc=1pt,
  colback=#3, colframe=white,
  colbacktitle=#1, coltitle=white,
  fonttitle=\bfseries\footnotesize,
  attach boxed title to top left={xshift=2mm, yshift=-2mm},
  boxed title style={arc=1pt, boxrule=0pt},
  title=#2, top=4mm, before skip=2pt, after skip=2pt}

\newenvironment{role}{\begin{promptsec}{rolecol}{(i) Role}{white}}{\end{promptsec}}
\newenvironment{actions}{\begin{promptsec}{actcol}{(ii) Valid Actions}{white}}{\end{promptsec}}
\newenvironment{rules}{\begin{promptsec}{rulecol}{(iii) Rules}{white}}{\end{promptsec}}
\newenvironment{statebasic}{\begin{promptsec}{statecol}{(iv) State}{white}}{\end{promptsec}}
\newenvironment{stateenrich}{\begin{promptsec}{statecol}{(iv) State + enriched augmentation}{enrichbg}}{\end{promptsec}}
\newenvironment{statestateful}{\begin{promptsec}{statecol}{(iv) State + stateful augmentation}{statebg}}{\end{promptsec}}

\newenvironment{examples}{\begin{promptsec}{exmpcol}{(v) Examples}{white}}{\end{promptsec}}
\newenvironment{outputfmt}{\begin{promptsec}{outcol}{(vi) Output Format}{white}}{\end{promptsec}}

\usepackage[subpreambles=true]{standalone}
\usepackage{booktabs}
\usepackage{multirow}
\usepackage{mystyle}
\usepackage[inline]{enumitem}
\usepackage{algorithm}
\usepackage{algpseudocode}
\usepackage{tabularx}
\usepackage[
    textsize=scriptsize,
]{todonotes}
\usepackage{soul}

\newcommand{\abbrev}{ASK+}
\newcommand{\thinkbudget}{10} 
\newcommand{\anonimity}{\url{https://github.com/jrzmnt/ask-pomdp}}

\begin{document}

\copyrightyear{2026}
\copyrightclause{Copyright for this paper by its authors.
  Use permitted under Creative Commons License Attribution 4.0
  International (CC BY 4.0).}

\conference{PRL+CAIPI Workshop @ IJCAI-ECAI 2026}

\title{ASK in the Dark: Uncertainty-Gated LLM Assistance under Partial Observability}

\author[1]{Juarez Monteiro}[%
email=juarez@kunumi.com,
]
\cormark[1]
\fnmark[1]
\address[1]{Kunumi Institute, Belo Horizonte, MG, Brazil}

\begingroup\catcode`\_=12
\author[2]{Nathan Gavenski}[%
email=nathan.schneider_gavenski@kcl.ac.uk,
]
\endgroup
\fnmark[1]
\address[2]{King's College London, London, United Kingdom}

\author[1]{Guilherme Lima}[%
email=adriano@kunumi.com,
]

\author[1]{Francisco Galuppo}[%
email=francisco@kunumi.com
]

\author[2]{Odinaldo Rodrigues}[%
email=odinaldo.rodrigues@kcl.ac.uk
]

\author[1]{Adriano Veloso}[%
email=adriano@kunumi.com,
]
\cortext[1]{Corresponding author.}
\fntext[1]{These authors contributed equally.}

\begin{abstract}
Reinforcement learning agents operating under partial observability must act on incomplete information, making them natural candidates for guidance from small language models (SLMs) that carry broad reasoning priors. Yet integrating SLM guidance into this setting has proven difficult: across all test environments, vanilla uncertainty-gated approaches achieve an overwrite rate at or near zero,  meaning the SLM almost never contributes an independent action.
We trace this failure to the bare egocentric prompt, which provides insufficient context for genuine reasoning, and identify it as a context problem rather than a capacity problem.
We propose \abbrev{}, which supplies the SLM with trajectory-aware context (a partially revealed map, visited positions, and action history) and structured chain-of-thought reasoning, converting it from a passive redundancy check into a more informative consultant that occasionally corrects the policy.
We further establish that the predictive entropy signal used for selective querying measures action uncertainty rather than state uncertainty and remains informative in POMDPs, making uncertainty-gated assistance viable beyond fully observable settings.
The stateful prompt drives substantial gains: on DoorKey, where vanilla ASK matches PPO (both $89\%$), \abbrev{} reaches $\mathbf{93\%}$ success; on FourRooms, success climbs from $53\%$ to $\mathbf{70\%}$; on HigherLower, accuracy reaches $\mathbf{73.7\%}$, matching the SLM-only upper bound.
Across all environments, Qwen3.5-2B matches or exceeds Qwen3.5-4B, confirming that prompt design and selective gating dominate the impact of model scale, enabling guidance without large models.
\end{abstract}


\maketitle

\input{paper/01_introduction}
\input{paper/02_background}
\input{paper/03_method}
\input{paper/04_experiments}
\input{paper/05_results}

\input{paper/06_ablations}
\input{paper/08_related_work}
\input{paper/09_conclusion}

\begin{acknowledgments}
This work was partially supported by UK Research and Innovation [grant number EP/S023356/1], in the UKRI Centre for Doctoral Training in Safe and Trusted Artificial Intelligence, and by the Kunumi Institute, through individual grants awarded to the authors.
\end{acknowledgments}

\section*{Declaration on Generative AI}
During the preparation of this work, the authors used Claude (Anthropic), Grammarly in order to: improve writing style, grammar and spelling check.
After using these tools, the authors reviewed and edited the content as needed and took full responsibility for the publication’s content.

\bibliography{references}

\clearpage

\include{paper/10_apendix}

\end{document}

%% file: paper/01_introduction.tex
\section{Introduction}
\label{sec:intro}


Agents operating under partial observability must act on incomplete information, relying on egocentric observations that may be insufficient to distinguish between meaningfully different underlying states.
This setting is particularly challenging for reinforcement learning (RL) agents, which learn during training to approximate a value function that reflects the specific structure of the training environment.
When that structure changes, such as the repositioning of a door, the agent has no mechanism to detect the mismatch.
Its observation looks familiar, and it follows its value function confidently toward failure.

Recent work~\cite{monteiro2026ask,ahn2022saycan} addresses this challenge by combining external guidance from language models (LMs) with the RL agent.
LMs are a natural candidate for this role: trained on vast amounts of human knowledge, they can reason about goals, infer structure, and suggest sensible actions even in novel situations.
However, this reasoning ability is contingent on having sufficient context.
Under partial observability, the agent's view of the world is inherently limited, and an LM consulting only a narrow, egocentric observation may lack the global context needed to reason reliably, producing guidance that is confidently wrong rather than helpfully corrective~\cite{valmeekam2025systematic}.

The ASK framework~\cite{monteiro2026ask} addresses this by selectively gating small LM (SLM) queries, consulting the language model only when the policy's uncertainty exceeds a threshold.
In its original evaluation on FrozenLake, a fully observable environment, this selective strategy allowed ASK to approach PPO-level performance while keeping the number of model queries low.
This suggests that a limited reasoning budget, spent wisely, can compensate for policy uncertainty without overwhelming it.
Whether this gating mechanism remains effective under partial observability is a question the original work leaves open, and the one this paper sets out to answer.



We evaluate an extension of ASK, which we coin \abbrev{}, across three partially observable environments that stress-test its gating mechanism in complementary ways.
\texttt{MiniGrid-FourRooms-v0}~\cite{minigrid} requires the agent to navigate a 19$\times$19 grid from an egocentric 7$\times$7 view to reach a sparse goal, isolating the challenge of spatial grounding under limited visibility.
\texttt{MiniGrid-DoorKey-8x8}~\cite{minigrid} extends this with a sequential subtask structure requiring the agent to find a key, unlock a door, and reach the goal.
\texttt{POPGym-HigherLower}~\cite{morad2023popgym} requires the agent to predict card ranks from a single observation per step, where optimal play depends on tracking all previously seen cards.
Together, the three environments isolate distinct failure modes of language-model assistance under partial observability: spatial grounding, sequential task decomposition, and temporal memory.
Our contributions are as follows.
\begin{enumerate}
    \item We propose \abbrev{}, which diagnoses vanilla ASK's failure under partial observability as a context bottleneck rather than a capacity limitation, and addresses it with stateful, episodic prompting that supplies the SLM with a partially revealed map, visited positions, action history, and structured chain-of-thought reasoning. We show that the underlying entropy signal remains informative under partial observability, and that \abbrev{} consistently outperforms both vanilla ASK and the PPO baseline across three complementary partially observable Markov decision processes (POMDPs).
    \item We provide an empirical comparison showing that prompt design and selective gating, rather than model scale, are the primary drivers of performance, with Qwen3.5-2B matching or exceeding Qwen3.5-4B throughout our experiments.
\end{enumerate}


%% file: paper/02_background.tex
\section{Background}
\label{sec:background}

We formulate the partial observability problem as a Partially Observable Markov Decision Process (POMDP)~\cite{kaelbling1998planning}. 
A POMDP is a tuple $\mathcal{M} = \langle S, A, T, r, \gamma, O,  \Omega \rangle$, where $S$ is the state-space set, $A$ the action-space set, $T: S \times A \rightarrow S$ is the transition function, $r: S \times A \rightarrow \mathbb{R}$ is the expected immediate reward for taking action $a$ in state $s$, $\gamma \in [0, 1)$ is the discount factor, $O$ is the observation space set, and $\Omega: S \rightarrow O$ is the observation function.
Note that, unlike traditional Markovian processes, in POMDPs, the agent observes the observation produced by $\Omega$.
Solving the POMDP yields a policy $\pi: O \rightarrow A$ that maps observations to actions so as to maximize the expected discounted return $\mathbb{E}\left[\sum_{t=0}^{\infty} \gamma^t r(s_t, a_t)\right]$.
We treat observations as incomplete state representations: $\Omega$ is non-injective, so distinct states may produce identical observations and the agent cannot recover $s$ from $o$ alone.

We use the ASK~\cite{monteiro2026ask} framework to leverage both the RL policy and the SLM model.
ASK is an extrinsic method that augments a trained RL policy with an SLM consultant without retraining or architectural modification.
At each step, ASK applies MC Dropout~\cite{gal2016dropout} to the policy's MLP (multi-layer perceptron): it executes $N$ stochastic forward passes with dropout active, producing a distribution over action probabilities whose entropy serves as the uncertainty signal.
If the total predictive entropy $H(o)$ meets or exceeds a threshold $\tau$, the SLM is queried with the current observation (and the PPO suggestion, for ASK); its response overrides the policy if valid.
The threshold $\tau$ is selected via Bayesian optimization on a held-out validation set; the test split is never accessed during tuning.
ASK was originally evaluated on FrozenLake, a discrete, fully observable grid, where selective querying achieved robust out-of-distribution generalization at low intervention rates.
We investigate whether the same entropy-based signal remains a reliable trigger when the policy operates on ambiguous partial observations, and whether SLM's foundational knowledge is sufficient to solve partially observable tasks.

%% file: paper/03_method.tex
\section{Method}
\label{sec:method}

\begin{figure}[t]
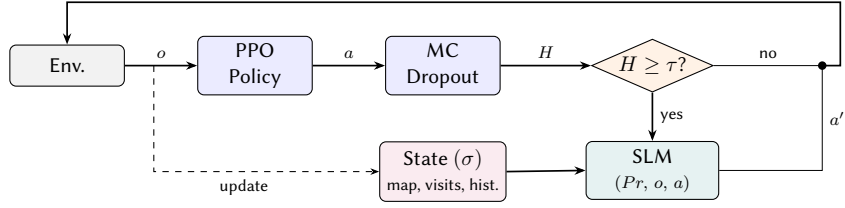

    \centering
    \includestandalone[width=.7\linewidth]{figures/ask_diagram}
    \caption{Decision flow of \abbrev{}. At each step, the PPO policy produces a default action $a$, and MC Dropout estimates the total predictive entropy $H(o)$. If $H(o) \geq \tau$, the SLM is queried via a prompt $Pr$ constructed from the current observation, the PPO suggestion, and the episode state $\sigma$; its output $a'$ overrides $a$ when valid. The episode state $\sigma$ (dashed arrow) accumulates observations, visited positions, and action history throughout the episode, providing global context that is absent from any single egocentric observation.}
    \label{fig:ask-diagram}
\end{figure}

\abbrev{} extends ASK~\cite{monteiro2026ask} to partially observable settings by augmenting its uncertainty-gated querying mechanism with stateful episode reasoning (Figure~\ref{fig:ask-diagram}): MC Dropout estimates the total predictive entropy $H(o)$ via $N$ stochastic forward passes; if $H(o) < \tau$ the PPO action executes directly and $\sigma$ is updated, otherwise it queries the SLM via a prompt $Pr$ built from $o$, $a$, and $\sigma$.

The principal departure from ASK is in how \abbrev{} elicits the SLM's reasoning capabilities.
The original ASK supplies only the current observation, depriving the model of the context it needs to exercise its foundational knowledge: a language model restricted to the agent's immediate field of view cannot reason spatially, track history, or justify its suggestions.
\abbrev{} addresses this by designing the prompt interface to unlock three complementary reasoning capacities.
The prompt is \emph{enriched} with derived environmental features and local action previews, enabling the SLM to perform spatial inference rather than reactive pattern matching on raw observations.
It is \emph{stateful}, appending episodic memory built from $\sigma$, i.e., a partially revealed map, visited positions, and action history, so that the SLM can reason over the trajectory rather than an isolated snapshot.
Finally, it activates the SLM's native thinking mode within a constrained budget, giving the model a bounded window of deliberate computation before it commits to an action, rather than forcing an immediate one-word reaction.

\subsection{Language Model Assistance}\label{sec:sub:lm-assistance}

\abbrev{} targets SLMs rather than frontier models, extending ASK's original reliance on larger systems with models that are computationally viable for step-level inference.
The rationale for this choice is that recent SLMs encode sufficient world knowledge and can perform structured reasoning~\cite{qwen3.5}.
Yet this capacity is latent and requires the right context to be activated.
Supplying a single egocentric observation, as ASK does, provides neither the spatial grounding nor the sequential context that these models need to engage their reasoning mechanisms, resulting in insufficiently informed predictions.

Therefore, \abbrev{} treats the prompt as an interface to the model's reasoning rather than a query format, and systematically varies the amount of context provided through three configurations.
The \emph{enriched} configuration extends the raw observation with derived environmental features and local action previews, giving the model the inputs it needs for spatial inference.
The \emph{stateful} configuration additionally injects episodic memory from $\sigma$, enabling the model to condition on trajectory history and reason about what regions have been explored or which moves have been attempted.
The \emph{rationale} configuration further enables the SLM's native thinking mode, allocating a constrained budget of \thinkbudget{} tokens for deliberate computation before committing to action, without incurring the computational overhead that would undermine the efficiency case for small models.
The visible output still closes with a short justification line. 
Still, this line is a gloss on the decision rather than a verified trace of the thinking process itself, and we do not treat it as evidence of reasoning beyond the mechanism just described.
These configurations form a progression: each adds a layer of context that makes a qualitatively different reasoning capacity accessible, rather than varying prompt style as a surface-level engineering choice.
We present a thorough ablation study that explores each configuration in Section~\ref{sec:sub:prompt}.

\subsection{Policy Architecture and Uncertainty Estimation}\label{sec:sub:uncertainty}

The RL policy is a standard actor-critic whose neural network extractor incorporates Monte Carlo Dropout, following the uncertainty decomposition described in Section~\ref{sec:background}.
Dropout is retained at inference time to produce a distribution of action estimates across $N$ stochastic passes, from which total predictive entropy $H(o)$, aleatoric uncertainty, and epistemic uncertainty are derived.
The gating threshold is applied to $H(o)$ rather than the epistemic component alone, as both sources of uncertainty indicate that the policy's action estimate is unreliable: aleatoric uncertainty reflects genuine observation ambiguity, while epistemic uncertainty reflects insufficient training coverage.
Although most works focus on epistemic uncertainty, we advocate for total uncertainty, since aleatoric uncertainty, while irreducible from the RL policy perspective, may still be solvable by leveraging the foundational knowledge and reasoning capabilities from SLMs.

A key methodological implication of the hybrid architecture is that the RL policy no longer needs to generalize to high-uncertainty states.
Generalization under partial observability is typically expensive to achieve through RL alone: the policy must implicitly learn to track history and resolve ambiguity from the reward signal alone, which requires high parametric capacity and extended training.
In \abbrev{}, those states are delegated to the SLM, whose pretraining has already instilled the broad world knowledge and reasoning structure that RL cannot easily acquire.
This allows the PPO policy to remain small and focused on in-distribution states, where it is both efficient and accurate.
The SLM's overparametrisation relative to the immediate task is not a liability; it is the source of the generalization capacity the policy intentionally forgoes.
The threshold $\tau$ then functions as a capacity allocation mechanism, determining how much of the SLM's representational surplus is recruited to compensate for the policy's constrained generalization.
Nevertheless, we hypothesize that SLMs cannot solve these tasks on their own and require the RL policies' local, task-specific knowledge to succeed.
To understand the impact of the RL policy in \abbrev{}, we experiment with different degrees of sub-optimality in Section~\ref{sec:sub:suboptimal}.


%% file: figures/ask_diagram.tex
\begin{tikzpicture}[
    box/.style={draw, rounded corners=3pt, minimum width=1.9cm,
                minimum height=0.68cm, align=center,
                font=\small, line width=0.5pt},
    dia/.style={draw, diamond, aspect=2.3, inner sep=1pt,
                align=center, font=\small, line width=0.5pt},
    arr/.style={-{Stealth[scale=0.8]}, line width=0.8pt},
    upd/.style={-{Stealth[scale=0.8]}, line width=0.5pt, dashed},
    lbl/.style={font=\scriptsize},
]

\node[box, fill=gray!10]                (env)   {Env.};
\node[box, fill=blue!8,  right=1.2cm of env]    (ppo)   {PPO\\Policy};
\node[box, fill=blue!8,  right=1.2cm of ppo]    (mc)    {MC\\Dropout};
\node[dia, fill=orange!10, right=1.5cm of mc]   (gate)  {$H \geq \tau$?};

\node[box, fill=purple!8, below=0.76cm of mc,
      minimum width=1.9cm, minimum height=0.9cm] (state)
      {State~$(\sigma)$\\{\scriptsize map, visits, hist.}};
\node[box, fill=teal!10,  below=0.8cm of gate,
      minimum width=2.2cm, minimum height=0.9cm] (slm)
      {\textsc{SLM}\\{\scriptsize$(Pr,\,o,\,a)$}};

\coordinate (R) at ($(gate.east)+(1.8cm,0)$);
\coordinate (obranch) at ($(env.east)!0.4!(ppo.west)$);

\draw[arr] (env.east)   -- node[above,lbl]{$o$}  (ppo.west);
\draw[arr] (ppo.east)   -- node[above,lbl]{$a$}  (mc.west);
\draw[arr] (mc.east)    -- node[above,lbl]{$H$}  (gate.west);

\draw[arr] (gate.south) -- node[right,lbl]{yes}  (slm.north);

\draw[arr] (state.east) --                        (slm.west);

\draw[upd] (obranch) |- node[below,lbl, pos=0.7]{update} (state.west);

\draw      (gate.east)  -- node[above,lbl]{no}   (R);
\draw      (slm.east)   -| node[right,lbl,near end]{$a'$} (R);
\fill      (R) circle (2pt);
\draw[arr] (R) -- ++(0.3,0) |- ($(env.north)+(0,0.7cm)$) -- (env.north);

\end{tikzpicture}

%% file: paper/04_experiments.tex
\section{Experiments and Metrics}
\label{sec:experiments}

In this work, we use Stable-Baselines3's PPO implementation~\cite{stable_baselines3} as the base policy, trained for $2\times 10^6$ steps with default hyperparameters and a linear learning rate schedule.
As for the SLM, we use the $2$ and $4$ billion parameter variants of the Qwen $3.5$ family~\cite{qwen3.5}, which are publicly available and have demonstrated strong foundational knowledge and reasoning capabilities~\cite{gavenski2026doubtplanoutcommitted}.
No fine-tuning is performed on the SLM, and the PPO model is trained independently without any SLM interaction, ensuring a clean evaluation of the ASK framework under partial observability.
All agents are evaluated on the same $100$ test episodes (seeds $[100, 199]$), with the validation split (seeds $[0, 99]$) reserved exclusively for threshold tuning.
The official \abbrev{} codebase is available at \emph{\anonimity}.

As testbeds, we use the FourRooms~\cite{minigrid}, DoorKey~\cite{minigrid}, and HigherLower~\cite{morad2023popgym} environments, which are designed to stress spatial grounding and memory-based reasoning.
Minigrid's DoorKey and FourRooms are procedurally generated gridworlds with sparse rewards, while POPGym-HigherLower is a card game where optimal play requires tracking the full deck history from a single-rank observation.
We select these environments since they are standard benchmarks in the literature.
Moreover, they present complementary challenges that allow us to evaluate the \abbrev{} under varying degrees of partial observability and reasoning complexity.
We select the gating threshold $\tau \in [0.1,\,2.0]$ independently for each SLM-environment pair via Optuna~\cite{optuna}, which uses the Tree-structured Parzen Estimator algorithm.

We report the reward, success (for DoorKey and FourRooms) or accuracy (for HigherLower), intervention rate (IR), and overwrite rate (OR) for each baseline and \abbrev{} configuration.
Reward is the accumulated return per episode, success/accuracy is the fraction of successful episodes (DoorKey and FourRooms) or correct decisions (HigherLower), IR is the fraction of steps where the SLM is queried, and OR is the fraction of steps where the SLM's action differs from the PPO's action.
IR and OR together characterize the efficiency--performance trade-off: IR quantifies computational overhead; OR measures how often the SLM provides guidance beyond validating the policy's own choice.
We compare against three baselines: (i) PPO-only, which never queries the SLM; (ii) SLM-only, which always queries the SLM and ignores the PPO; and (iii) ASK, with the original prompting strategy.


%% file: paper/05_results.tex
\section{Results}
\label{sec:results}

\input{tables/results}

Table~\ref{tab:main} reports mean reward, success/accuracy, IR, and OR across all three environments over 100 test episodes.
\abbrev{} achieves the highest reward and success in both navigation environments, FourRooms and DoorKey-8$\times$8, while matching or narrowly exceeding the baselines in HigherLower.
SLM-only, without the RL backbone, degrades substantially on DoorKey, confirming that the RL policy's task-specific knowledge remains indispensable even when the language model is given complete control. On FourRooms and HigherLower, however, SLM-only matches or approaches \abbrev{} on raw performance. In these settings \abbrev{}'s contribution is efficiency rather than peak quality.

The most consistent gains from \abbrev{} are observed in navigation environments, where spatial grounding is the dominant challenge.
In FourRooms, \abbrev{}-$2$B raises reward from $0.50$ (PPO) and $0.51$--$0.52$ (ASK) to $0.64$, a difference of $0.12$--$0.14$ reward points, with a corresponding increase in success rate from $0.53$--$0.55$ to $0.70$.
DoorKey shows a similar pattern: \abbrev{}-$2$B reaches a success rate of $0.93$, improving over ASK-$4$B ($0.92$) and PPO ($0.89$).
Note that ASK's OR is at or near zero ($0.00$--$0.01$) across all environments, meaning the SLM is queried but never produces an action that differs from the PPO's suggestion, effectively functioning as a passive validator rather than an active corrector.
By contrast, \abbrev{} yields nonzero ORs ($0.01$--$0.02$ in the navigation tasks), indicating that the enriched, stateful prompt elicits genuinely informative responses rather than mere policy confirmation.

HigherLower presents a different picture.
The environment rewards accurate card prediction, and all methods cluster within a narrow performance band: reward ranges from $0.50$ to $0.53$ and accuracy from $0.72$ to $0.74$.
\abbrev{} improves accuracy from $0.72$--$0.73$ (PPO and ASK) to $0.74$, matching SLM-$4$B, but does not surpass it outright.
The IR in HigherLower is notably higher than in the navigation tasks, reaching $0.60$--$0.69$ for both ASK and \abbrev{}, reflecting that the entropy-gating fires more frequently in this memory-dependent setting where individual observations are inherently ambiguous.
While \abbrev{} achieves nonzero ORs here as well ($0.05$--$0.06$), the marginal performance gap suggests that the memory component of the prompt, though activated, is not sufficient to overcome the SLM's advantage of observing the full card history at every step.

Nevertheless, interpreting IR and OR jointly requires some care.
A high IR is not inherently undesirable.
The gating mechanism is designed to query the SLM precisely when the policy is uncertain, and doing so at those steps is the intended behavior regardless of whether the SLM ultimately changes the action.
Conversely, an overwrite is not a reliable proxy for quality.
That is, the SLM overriding the policy more frequently does not imply that the resulting trajectory is better.
A query that confirms the policy's predicted action under uncertainty still provides a meaningful check, and the performance gains reported here demonstrate that this can translate into measurable improvements over the baselines, even when OR remains low.
What ASK's near-zero OR reveals is not that its consultations are valuable confirmations, but rather that the bare egocentric prompt provides insufficient context for the SLM to form any independent judgment, leaving it unable to contribute beyond echoing the policy's suggestion.

These results show that \abbrev{}'s stateful and enriched prompting converts the SLM from a passive redundancy check into a more informative consultant, one that corrects the policy on a small but consistent fraction of steps, and that this conversion translates to meaningful performance gains in environments where spatial reasoning benefits from historical context.
The narrower improvement in HigherLower suggests that the gains are most reliable when the SLM's reasoning is grounded in navigational context, rather than when the task reduces to sequence memory, which the SLM-only baseline already handles well.

%% file: tables/results.tex
\begin{table*}[t]
\setlength{\tabcolsep}{1pt}
\scriptsize
\caption{%
  Main results across all three environments.
  IR = 0 for PPO and 1.00 for SLM-only by design.
}
\label{tab:main}
\centering
\begin{tabular*}{\textwidth}{@{\quad}ll@{\extracolsep{\fill}} cccc cccc cccc@{\quad}}
\toprule
& & \multicolumn{4}{c}{\textbf{FourRooms}}
  & \multicolumn{4}{c}{\textbf{HigherLower}}
  & \multicolumn{4}{c}{\textbf{DoorKey-8$\times$8}} \\[-2pt]
\cmidrule(lr){3-6}\cmidrule(lr){7-10}\cmidrule(lr){11-14}
\textbf{Agent} & \textbf{Model}
  & \textbf{Reward}~$\uparrow$  & \textbf{Succ}~$\uparrow$  & \textbf{IR}~$\downarrow$  & \textbf{OR}~$\downarrow$
  & \textbf{Reward}~$\uparrow$  & \textbf{Acc}~$\uparrow$   & \textbf{IR}~$\downarrow$  & \textbf{OR}~$\downarrow$
  & \textbf{Reward}~$\uparrow$  & \textbf{Succ}~$\uparrow$  & \textbf{IR}~$\downarrow$  & \textbf{OR}~$\downarrow$ \\[-2pt]
\midrule
PPO      & ---
  & $0.50 \pm 0.48$ & $0.53$ & $0.00$ & ---
  & $0.50 \pm 0.11$ & $0.72$ & $0.00$ & ---
  & $0.87 \pm 0.31$ & $0.89$ & $0.00$ & --- \\[-2pt]
\midrule
SLM & 2B
  & $0.30 \pm 0.43$ & $0.34$ & $1.00$ & ---
  & $0.51 \pm 0.10$ & $0.73$ & $1.00$ & ---
  & $0.00 \pm 0.00$ & $0.00$ & $1.00$ & --- \\[-2pt]
SLM & 4B
  & $0.57 \pm 0.42$ & $0.69$ & $1.00$ & ---
  & $\mathbf{0.53 \pm 0.10}$ & $\mathbf{0.74}$ & $1.00$ & ---
  & $0.58 \pm 0.46$ & $0.62$ & $1.00$ & --- \\[-2pt]
\midrule
ASK      & 2B
  & $0.51 \pm 0.47$ & $0.54$ & $0.32$ & $0.00$
  & $0.50 \pm 0.11$ & $0.73$ & $0.69$ & $0.00$
  & $0.87 \pm 0.31$ & $0.89$ & $0.06$ & $0.01$ \\[-2pt]
ASK      & 4B
  & $0.52 \pm 0.47$ & $0.55$ & $0.32$ & $0.00$
  & $0.50 \pm 0.11$ & $0.72$ & $\mathbf{0.60}$ & $0.00$
  & $0.89 \pm 0.27$ & $0.92$ & $0.05$ & $0.00$ \\[-2pt]
\midrule
\abbrev~(ours) & 2B
  & $\mathbf{0.64 \pm 0.43}$ & $\mathbf{0.70}$ & $\mathbf{0.20}$ & $0.02$
  & $0.52 \pm 0.11$          & $\mathbf{0.74}$ & $0.69$          & $0.06$
  & $\mathbf{0.91 \pm 0.25}$ & $\mathbf{0.93}$ & $\mathbf{0.03}$ & $0.02$ \\[-2pt]
\abbrev~(ours) & 4B
  & $0.62 \pm 0.43$          & $0.69$          & $0.25$          & $0.02$
  & $0.52 \pm 0.10$          & $\mathbf{0.74}$ & $\mathbf{0.60}$ & $0.05$
  & $0.90 \pm 0.25$          & $\mathbf{0.93}$ & $\mathbf{0.03}$ & $0.02$ \\[-2pt]
\bottomrule
\end{tabular*}
\end{table*}

%% file: paper/06_ablations.tex
\section{Ablations}
We run four ablations to isolate the contribution of each design choice.
First, we test the impact of prompt design by comparing the basic strategy (resembling ASK's) against the enriched, stateful, and rationale-augmented variants.
Second, we perform a grid search over the uncertainty threshold $\tau$ to confirm that it effectively controls the intervention rate and reward tradeoff.
Finally, we evaluate \abbrev{} with suboptimal PPO checkpoints to test its robustness to a weak base policy.

\subsection{Prompt Components Contribution}\label{sec:sub:prompt}

\begin{table}[b]
\centering
\scriptsize
\setlength{\tabcolsep}{1pt}
\caption{%
    \abbrev{} prompt ablation across the three environments (we keep $\tau$ fixed accross prompt strategies for consistency).
}
\label{tab:askprompt-all}
\begin{tabular*}{\textwidth}{@{\quad}ll@{\extracolsep{\fill}} cccc cccc@{\quad}}
\toprule
\multirow{2}{*}{\textbf{Env}} & \multirow{2}{*}{\textbf{Prompt}}
 & \multicolumn{4}{c}{\textbf{\abbrev{} --- Qwen3.5-2B}}
 & \multicolumn{4}{c}{\textbf{\abbrev{} --- Qwen3.5-4B}} \\[-2pt]
\cmidrule(lr){3-6} \cmidrule(lr){7-10}
 &
 & Rew.\ $\uparrow$ & Succ./Acc.\ $\uparrow$ & IR $\downarrow$ & OR $\downarrow$
 & Rew.\ $\uparrow$ & Succ./Acc.\ $\uparrow$ & IR $\downarrow$ & OR $\downarrow$ \\[-2pt]
\midrule
\multirow{5}{*}{\textbf{FourRooms}}
 & basic                 & $0.513$ & $0.54$ & $0.32$ & $0.00$
                         & $0.523$ & $0.55$ & $0.32$ & $0.00$ \\[-2pt]
 & enriched              & $0.523$ & $0.55$ & $0.31$ & $0.00$
                         & $0.522$ & $0.55$ & $0.31$ & $0.00$ \\[-2pt]
 & stateful\_min         & $0.523$ & $0.55$ & $0.31$ & $0.00$
                         & $0.523$ & $0.55$ & $0.32$ & $0.00$ \\[-2pt]
 & stateful              & $0.522$ & $0.55$ & $0.31$ & $0.00$
                         & $0.523$ & $0.55$ & $0.32$ & $0.00$ \\[-2pt]
 & stateful + rationale  & $\mathbf{0.640}$ & $\mathbf{0.70}$ & $0.20$ & $0.02$
                         & $\mathbf{0.621}$ & $\mathbf{0.69}$ & $0.25$ & $0.02$ \\[-2pt]
\midrule
\multirow{4}{*}{\textbf{HigherLower}}
 & basic                 & $0.501$ & $0.726$ & $0.69$ & $0.00$
                         & $0.495$ & $0.723$ & $0.60$ & $0.00$ \\[-2pt]
 & enriched              & $\mathbf{0.525}$ & $\mathbf{0.738}$ & $0.69$ & $0.06$
                         & $\mathbf{0.524}$ & $\mathbf{0.738}$ & $0.60$ & $0.06$ \\[-2pt]
 & stateful              & $0.515$ & $0.733$ & $0.69$ & $0.03$
                         & $0.516$ & $0.734$ & $0.60$ & $0.03$ \\[-2pt]
 & stateful + rationale  & $0.521$ & $0.736$ & $0.69$ & $0.06$
                         & $0.520$ & $0.736$ & $0.60$ & $0.05$ \\[-2pt]
\midrule
\multirow{5}{*}{\textbf{DoorKey$_{8\!\times\!8}$}}
 & basic                 & $0.869$ & $0.89$ & $0.06$ & $0.01$
                         & $0.894$ & $0.92$ & $0.05$ & $0.00$ \\[-2pt]
 & enriched              & $0.859$ & $0.88$ & $0.05$ & $0.03$
                         & $0.843$ & $0.87$ & $0.06$ & $0.05$ \\[-2pt]
 & stateful\_min         & $\mathbf{0.917}$ & $\mathbf{0.94}$ & $0.03$ & $0.01$
                         & $0.877$ & $0.91$ & $0.05$ & $0.04$ \\[-2pt]
 & stateful              & $0.908$ & $0.93$ & $0.03$ & $0.01$
                         & $\mathbf{0.905}$ & $\mathbf{0.93}$ & $0.03$ & $0.02$ \\[-2pt]
 & stateful + rationale  & $0.908$ & $0.93$ & $0.02$ & $0.01$
                         & $0.896$ & $0.92$ & $0.03$ & $0.02$ \\[-2pt]
\bottomrule
\end{tabular*}
\end{table}

To understand how each prompt component contributes to \abbrev's performance, we design a systematic ablation of the prompt structure.
The basic prompt, which resembles that of ASK, provides only the current observation and a few examples, while the enriched component adds derived features and action previews, the stateful component appends episodic memory and accumulated state information, and the rationale component enables the SLM's native thinking mode, with a constrained budget of \thinkbudget{} tokens, before the SLM outputs the action, increasing the decision budget available to the model.

All prompts follow a shared template consisting of:
\begin{enumerate*}[label=(\roman*)]
    \item a role declaration (e.g., ``\texttt{You are a helpful assistant...}'');
    \item a \texttt{VALID ACTIONS} block;
    \item a \texttt{RULES} block specifying behavioural constraints (e.g., ``you cannot move through walls'');
    \item a task-specific state description (e.g., \texttt{The agent is in room A and can see a key on the ground});
    \item a small few-shot \texttt{Examples} section illustrating the correct action for a given state; and
    \item a strict \texttt{OUTPUT FORMAT} requiring a single action token.
\end{enumerate*}
The three prompt components described in Section~\ref{sec:sub:lm-assistance} are implemented as augmentations to the state description section, which is the most semantically rich part of the prompt and therefore the most likely to activate the SLM's reasoning capabilities.
Additionally, we add a ``stateful\_min'' variant that includes only the episodic memory, without any map information.
Appendix~\ref{app:prompt} contains the full prompt templates for each environment and configuration.

Table~\ref{tab:askprompt-all} shows that prompt complexity interacts strongly with environment structure.
In FourRooms, the basic and enriched variants perform nearly identically (reward $\approx 0.52$, success rate $\approx 0.55$ for both model sizes), and adding episodic memory via the stateful and stateful\_min components produces no measurable gain.
However, the rationale component yields a pronounced jump: stateful + rationale raises reward to $0.640$ and success rate to $0.70$ for the 2B model, and to $0.621$ and $0.69$ for the 4B model, while simultaneously reducing the invalid-action rate.
The pattern reverses in HigherLower, where the enriched component alone achieves the highest accuracy ($0.738$ for both models), and both the stateful and stateful+rationale variants fall marginally short; the comparatively low invalid-action rate across all conditions suggests that this task is lexically unambiguous and benefits more from derived feature context than from extended reasoning.
DoorKey presents a third regime: the enriched component slightly underperforms the basic prompt, whereas the stateful\_min and stateful variants attain the best results, and the addition of the rationale component offers no further improvement, indicating that explicit state tracking is sufficient once the task structure is well-defined.

Taken together, these results demonstrate that no single prompt component is universally optimal across environments, and their relative contributions depend on the cognitive demands of each task.
Environments requiring spatial planning benefit from extended reasoning budgets, those with rich observable structure are best served by derived feature augmentation, and environments with complex multi-step dependencies rely on accumulated state information.
Omitting any one component risks a systematic deficit in at least one setting; the full prompt design provides complementary scaffolding that covers the range of demands across environments.

\subsection{Threshold $\tau$.}

We perform a grid search over $\tau \in [0.1, 2.0]$ in increments of $0.2$, bypassing Optuna, to confirm that it effectively controls the IR and reward tradeoff.
We note that for each environment, the optimal $\tau$ varies because it depends on the distribution of predictive entropies observed during episodes, which is influenced by the environment's complexity and the PPO policy's performance.

\begin{figure}[b]
  \centering
  \includegraphics[width=0.8\linewidth]{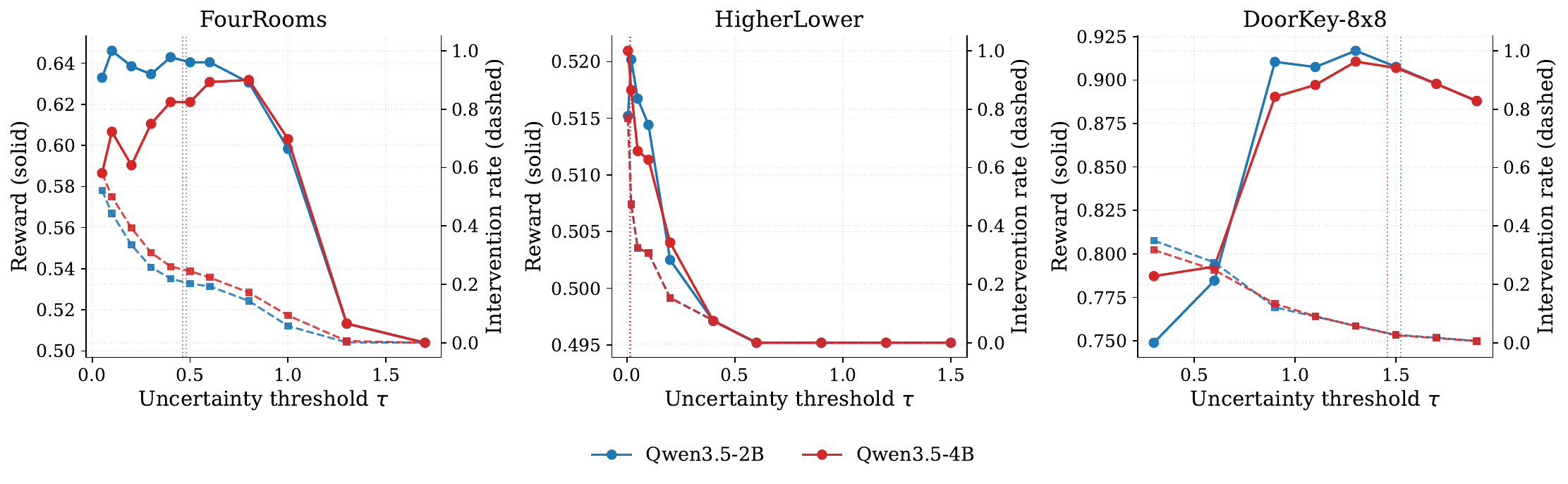}
  \caption{%
    Threshold $\tau$ for reward and IR effects, with the selected $\tau$ for each model marked with a line.
  }
  \label{fig:thresh}
\end{figure}

Figure \ref{fig:thresh} shows reward and IR across values of $\tau$. In FourRooms and DoorKey, low $\tau$ yields high IR and degraded reward, whereas in HigherLower it improves average reward. In all environments, high $\tau$ recovers PPO-level performance as IR approaches zero, since the SLM is bypassed entirely.
These results confirm that $\tau$ predictably controls the intervention rate, with its optimal value being environment-dependent rather than universal. 
The asymmetry between HigherLower and the navigation environments reflects differences in relative SLM--policy competence: in HigherLower, the SLM's numerical reasoning is sufficiently reliable that aggressive intervention improves outcomes, whereas in FourRooms and DoorKey, excessive intervention degrades reward because the SLM cannot substitute for the spatial planning encoded in the policy. 
The Optuna-selected values consistently fall in intermediate regimes that balance these pressures, neither bypassing the SLM nor over-delegating, validating automated threshold selection over manual tuning.

\subsection{PPO optimality}\label{sec:sub:suboptimal}

Finally, we test the robustness of SLM assistance to PPO suboptimality by reusing PPO checkpoints saved at different reward thresholds during training.
We expect that the SLM will be able to compensate for some degree of suboptimality, but that performance will degrade as the PPO policy becomes too weak to provide useful local guidance or to reach states where the SLM's reasoning can be effective.

\begin{table}[b]
\caption{
    Different results of \abbrev{} with different PPO checkpoints, which have different suboptimality levels.
}
\label{tab:ppoopt-all}
\centering
\scriptsize
\setlength{\tabcolsep}{1pt}
\begin{tabular*}{\textwidth}{@{\quad}ll@{\extracolsep{\fill}} cccc cccc cccc@{\quad}}
\toprule
\multirow{2}{*}{\textbf{Env}} & \multirow{2}{*}{\textbf{Ckpt}}
 & \multicolumn{4}{c}{\textbf{PPO}}
 & \multicolumn{4}{c}{\textbf{\abbrev{} --- Qwen3.5-2B}}
 & \multicolumn{4}{c}{\textbf{\abbrev{} --- Qwen3.5-4B}} \\[-2pt]
\cmidrule(lr){3-6} \cmidrule(lr){7-10} \cmidrule(lr){11-14}
 &
 & Rew.\ $\uparrow$ & Suc.\ $\uparrow$ & IR $\downarrow$ & $\tau$
 & Rew.\ $\uparrow$ & Suc.\ $\uparrow$ & IR $\downarrow$ & $\tau$
 & Rew.\ $\uparrow$ & Suc.\ $\uparrow$ & IR $\downarrow$ & $\tau$ \\[-2pt]
\midrule
\multirow{5}{*}{\textbf{FourRooms}}
 & $0.1$  & $0.18$ & $0.19$ & $0.0$ & ---
          & $0.34$ & $0.36$ & $0.55$ & $0.35$
          & $\mathbf{0.35}$ & $\mathbf{0.38}$ & $0.73$ & $0.18$ \\[-2pt]
 & $0.3$  & $0.20$ & $0.21$ & $0.0$ & ---
          & $0.38$ & $0.40$ & $0.41$ & $0.26$
          & $\mathbf{0.40}$ & $\mathbf{0.45}$ & $0.45$ & $0.26$ \\[-2pt]
 & $0.5$  & $0.45$ & $0.47$ & $0.0$ & ---
          & $\mathbf{0.57}$ & $\mathbf{0.62}$ & $0.45$ & $0.26$
          & $0.49$ & $0.53$ & $0.28$ & $0.47$ \\[-2pt]
 & $0.7$  & $0.60$ & $0.63$ & $0.0$ & ---
          & $0.65$ & $0.71$ & $0.37$ & $0.12$
          & $\mathbf{0.66}$ & $\mathbf{0.71}$ & $0.13$ & $0.86$ \\[-2pt]
 & full   & $0.50$ & $0.53$ & $0.0$ & ---
          & $\mathbf{0.64}$ & $\mathbf{0.70}$ & $0.20$ & $0.48$
          & $0.62$ & $0.69$ & $0.25$ & $0.46$ \\[-2pt]
\midrule
\multirow{5}{*}{\textbf{HigherLower}}
 & $0.1$  & $0.47$ & $0.71$ & $0.0$ & ---
          & $0.51$ & $0.73$ & $1.00$ & $0.42$
          & $\mathbf{0.52}$ & $\mathbf{0.74}$ & $1.00$ & $0.69$ \\[-2pt]
 & $0.2$  & $0.47$ & $0.71$ & $0.0$ & ---
          & $0.51$ & $0.73$ & $1.00$ & $0.55$
          & $\mathbf{0.52}$ & $\mathbf{0.74}$ & $1.00$ & $0.07$ \\[-2pt]
 & $0.3$  & $0.47$ & $0.71$ & $0.0$ & ---
          & $0.51$ & $0.73$ & $1.00$ & $0.13$
          & $\mathbf{0.52}$ & $\mathbf{0.74}$ & $1.00$ & $0.11$ \\[-2pt]
 & $0.4$  & $0.47$ & $0.71$ & $0.0$ & ---
          & $\mathbf{0.52}$ & $\mathbf{0.74}$ & $0.83$ & $0.91$
          & $0.52$ & $0.74$ & $1.00$ & $0.13$ \\[-2pt]
 & full   & $0.50$ & $0.72$ & $0.0$ & ---
          & $\mathbf{0.52}$ & $\mathbf{0.74}$ & $0.69$ & $0.01$
          & $0.52$ & $0.74$ & $0.60$ & $0.02$ \\[-2pt]
\midrule
\multirow{4}{*}{\textbf{DoorKey$_{8\!\times\!8}$}}
& $0.3$  & $0.29$ & $0.30$ & $0.0$ & ---
          & $0.50$ & $0.54$ & $0.52$ & $1.37$
          & $\mathbf{0.61}$ & $\mathbf{0.65}$ & $0.52$ & $0.37$ \\[-2pt]
 & $0.5$  & $0.65$ & $0.67$ & $0.0$ & ---
          & $0.76$ & $0.80$ & $0.30$ & $1.11$
          & $\mathbf{0.79}$ & $\mathbf{0.83}$ & $0.27$ & $0.19$ \\[-2pt]
 & $0.7$  & $0.72$ & $0.74$ & $0.0$ & ---
          & $0.78$ & $0.80$ & $0.13$ & $1.40$
          & $\mathbf{0.81}$ & $\mathbf{0.83}$ & $0.06$ & $0.03$ \\[-2pt]
 & full   & $0.87$ & $0.89$ & $0.0$ & ---
          & $\mathbf{0.91}$ & $\mathbf{0.93}$ & $0.03$ & $1.53$
          & $\mathbf{0.91}$ & $\mathbf{0.93}$ & $0.03$ & $1.46$ \\[-2pt]
\bottomrule
\end{tabular*}
\end{table}

Table~\ref{tab:ppoopt-all} shows that \abbrev{} consistently outperforms the corresponding PPO checkpoint at every level of policy suboptimality, and that the intervention rate scales inversely with checkpoint quality as expected.
In FourRooms, where the absolute gains are largest, \abbrev{} nearly doubles the reward of the weakest checkpoint (PPO $0.18$ versus \abbrev{} $0.34$--$0.35$ at the $0.1$ checkpoint), and IR progressively decreases from $0.55$--$0.73$ at the weakest checkpoint to $0.20$--$0.25$ at the full checkpoint, confirming that the entropy signal correctly identifies poorly calibrated policies.
DoorKey exhibits a similar pattern, with \abbrev{} recovering $0.50$-$0.61$ success from a PPO checkpoint at $0.30$ and maintaining an advantage at every subsequent level.
HigherLower is again distinctive: PPO reward plateaus early in training (all checkpoints hover around $0.47$--$0.50$), yet \abbrev{} reliably improves accuracy to $0.73$--$0.74$ across all checkpoints, with IR saturating at $1.00$ for the weakest policies, indicating that the system correctly recognizes the PPO signal as uninformative and delegates almost entirely to the SLM.

Taken together, these results demonstrate that \abbrev{} does not depend on a near-optimal PPO policy to deliver meaningful gains.
The uncertainty-gating mechanism adapts its intervention intensity to the quality of the underlying policy, intervening aggressively when the policy is unreliable and sparingly when it is competent, without requiring any retraining or recalibration.
This property is practically relevant because fully trained policies are not always available in deployment, and it suggests that \abbrev{} can function as a general-purpose corrective layer applied on top of policies at any stage of training.

%% file: paper/08_related_work.tex
\section{Related Work}
\label{sec:related}

Integrating language models into reinforcement learning has been explored primarily as a means of improving generalisation beyond training conditions.
\citet{ahn2022saycan} score LM-proposed actions through a learned affordance function, selecting the highest-scoring feasible action at each step; this approach is effective in constrained robotics settings but queries the LM unconditionally and offers no mechanism for deferring to the policy when the model is unreliable.
\citet{carta2023grounding} couple language planning with policy learning through online reinforcement learning, reducing the mismatch between language priors and environment dynamics, though the language component remains continuously active rather than invoked selectively.
ASK~\cite{monteiro2026ask} addresses this by gating LM queries on the policy's predictive entropy via MC Dropout~\cite{gal2016dropout}, achieving out-of-distribution generalisation without retraining the policy.
The motivation for selective rather than continuous consultation is reinforced by empirical evidence that LLMs struggle as autonomous sequential decision-makers~\cite{valmeekam2023planning, kambhampati2024llms, valmeekam2025systematic}, making indiscriminate querying likely to produce confidently incorrect guidance rather than useful correction.

Standard approaches to partial observability augment the policy with recurrent architectures or attention to accumulate history~\cite{kaelbling1998planning}; POPGym~\cite{morad2023popgym} provides a systematic benchmark of such memory requirements across diverse tasks.
Our work does not modify the policy architecture and instead asks whether an uncertainty-triggered language consultant can complement a memoryless policy in a POMDP, a setting that all prior LM-assisted RL work assumes away.
The three environments isolate distinct challenges: FourRooms requires egocentric spatial grounding, DoorKey introduces sequential subtasks, and HigherLower creates an asymmetric information setting where the SLM prompt exposes deck composition hidden from the policy.

%% file: paper/09_conclusion.tex

\section{Conclusion}
\label{sec:conclusion}

\abbrev{} extends ASK to partially observable settings by equipping the SLM with trajectory-aware context and structured chain-of-thought reasoning.
The context comprises a partially-revealed map, visited positions, and action history.
The key diagnostic that drives this design is that vanilla ASK achieves an overwrite rate at or near zero ($0.00$--$0.01$) across all three environments: the SLM is queried but almost never deviates from the policy's suggestion.
This exposes the bare egocentric prompt as the bottleneck rather than model capacity.
By closing this context gap, \abbrev{} provides the SLM with sufficient grounding to reason independently, lifting success from $53\%$ to $70\%$ in FourRooms, from $89\%$ to $93\%$ in DoorKey, and reaching $74\%$ accuracy in HigherLower.
The three environments were selected to isolate spatial grounding, sequential task decomposition and temporal memory as distinct failure modes of language-model assistance in the context of partial observability.
The gains confirm that the approach generalizes beyond the fully observable settings originally studied by ASK.

Taken together, these results show that uncertainty-gated SLM assistance is fundamentally sound under partial observability.
The failure of vanilla ASK is not a limitation of the gating idea but of the prompt, which gives the SLM too little context to act on.
The gating signal estimates total predictive entropy via MC Dropout, capturing both aleatoric uncertainty arising from genuine observational ambiguity and epistemic uncertainty due to insufficient training coverage.
This makes it effective whether the policy's unreliability stems from distributional shift or from the inherent state uncertainty of partial observability.
Qwen3.5-2B matches or exceeds Qwen3.5-4B across all environments, showing that the bottleneck is prompt design rather than model scale.
Ablations over $\tau$, $N$, PPO training quality, and prompting technique consistently identify selective querying and trajectory-aware context as the primary performance drivers, confirming that reasoning capacity can be leveraged without scaling up.

The present work intentionally scopes its evaluation to three complementary environments to establish a clean proof of concept.
Extending \abbrev{} to richer POMDPs, including settings in which the entropy signal misfires or stale history may mislead the SLM, is a natural next step.
Fine-tuning the SLM on egocentric views or task-specific trajectories would isolate the contribution of the gating mechanism from improved grounding.
Replacing the memoryless MLP with a recurrent policy in HigherLower would clarify how much of the \abbrev{} advantage stems from information asymmetry versus uncertainty gating.
Spatial analyses of where queries concentrate and ablations of the policy hint in the SLM prompt remain open to provide finer-grained mechanistic evidence.

%% file: paper/10_apendix.tex
\appendix

\section{Prompt Templates}\label{app:prompt}

All \abbrev{} prompts share a common six-block structure.
Block~(i) \emph{Role} and block~(ii) \emph{Valid Actions} are static per environment.
Block~(iii) \emph{Rules} encodes hard constraints, output format, and the rationale instruction when the rationale augmentation is active.
Block~(iv) \emph{State} carries the current observation and is the only block that changes at every step; the three augmentation layers (\emph{enriched}, \emph{stateful}, \emph{rationale}) each extend this block with progressively richer context.
Block~(v) \emph{Examples} provides one-shot demonstrations grounded in the environment's action vocabulary.
Block~(vi) \emph{Output Format} enforces the response schema.

The color coding used throughout this section is as follows.
{\footnotesize Header colors identify block type:
\colorbox{rolecol}{\color{white}\textbf{(i)\,Role}}\;
\colorbox{actcol}{\color{white}\textbf{(ii)\,Valid Actions}}\;
\colorbox{rulecol}{\color{white}\textbf{(iii)\,Rules}}\;
\colorbox{statecol}{\color{white}\textbf{(iv)\,State}}\;
\colorbox{exmpcol}{\color{white}\textbf{(v)\,Examples}}\;
\colorbox{outcol}{\color{white}\textbf{(vi)\,Output Format}}.
Background tints mark augmentation layers:
\fcolorbox{black}{enrichbg}{\textbf{enriched}}\;
\fcolorbox{black}{statebg}{\textbf{stateful}}\;
\fcolorbox{black}{ratbg}{\textbf{rationale}}.}

\input{prompts/four_rooms}
\input{prompts/higher_lower}
\input{prompts/door_key}

%% file: prompts/four_rooms.tex
\subsection{FourRooms}\label{app:prompt:fourrooms}

The FourRooms agent navigates a $19{\times}19$ grid from a $7{\times}7$ egocentric view.
The \emph{stateful} augmentation injects a partially-revealed map assembled from visited cells and a visit counter at the current position, giving the SLM a growing spatial record of explored territory.
The \emph{enriched} augmentation adds adjacency symbols (forward/left/right/back), ahead-chain depth, visible door locations, longest-corridor estimates in both the local view and the global map, and per-action previews showing the cell type that would be entered.
The \emph{rationale} block constrains the reasoning step to one line of at most 15 words, limiting inference overhead while still activating deliberative reasoning before action selection.

\begin{role}You are a robot navigation policy in MiniGrid FourRooms (partially observed). Choose exactly ONE low-level action.\end{role}

\begin{actions}\texttt{TURN\_LEFT \quad TURN\_RIGHT \quad FORWARD}\end{actions}

\begin{rules}
\begin{Verbatim}[fontsize=\scriptsize, baselinestretch=0.9, commandchars=\\\{\}]
- Output only the action token line required below (no markdown).
\colorbox{ratbg}{- Line 1: Reason: <one short line, max 15 words>.   Line 2: Action: <one of VALID ACTIONS>.}
- If FORWARD is blocked (ahead is '#'), do not choose FORWARD. Prefer doors (D)
  aligned with corridors / the cached goal. If 'G' is visible, move toward it.
- Follow planner hint and autopilot when consistent with safe previews.
\end{Verbatim}
\end{rules}

\begin{statestateful}
\begin{minipage}[t]{0.42\linewidth}
\begin{Verbatim}[fontsize=\scriptsize, baselinestretch=0.9]
CURRENT VIEW (egocentric):
.......
.....#.
.....#.
.....#.
.....#.
#######
???????
\end{Verbatim}
\end{minipage}\hfill
\begin{minipage}[t]{0.55\linewidth}
\begin{Verbatim}[fontsize=\scriptsize, baselinestretch=0.9]
DISCOVERED MAP (11x11 around A):
?????......
?????.#....
?????.#....
?????A#....
?????.#....
?????######
###########
\end{Verbatim}
\end{minipage}

\begin{Verbatim}[fontsize=\scriptsize, baselinestretch=0.9]
STATE: Facing=EAST  PathAhead=BLOCKED  Goal=not visible
       Pos=(8,16)   Room=SW            Visits@here=1   Actions=(none)
\end{Verbatim}
\end{statestateful}

\begin{stateenrich}
\begin{Verbatim}[fontsize=\scriptsize, baselinestretch=0.9]
Adjacent F/L/R/B: #/././.    Ahead chain: # . .    Doors visible: (none)
Longest ray (view): fwd 3    Longest ray (world): BACK 7    Autopilot: FORWARD
PREVIEW  TURN_LEFT->N,'.'    TURN_RIGHT->S,'.'    FORWARD->blocked '#'
\end{Verbatim}
\end{stateenrich}

\begin{examples}
\begin{Verbatim}[fontsize=\scriptsize, baselinestretch=0.9]
ahead=passable, goal visible (3 ahead)            -> FORWARD
ahead=BLOCKED,  door visible to the right         -> TURN_RIGHT
visits high + loop hint + longest_corridor=right  -> TURN_RIGHT
\end{Verbatim}
\end{examples}

\begin{outputfmt}
\begin{Verbatim}[fontsize=\scriptsize, baselinestretch=0.9, commandchars=\\\{\}]
\colorbox{ratbg}{Reason: <short>}   Action: TURN_LEFT or TURN_RIGHT or FORWARD
\end{Verbatim}
\end{outputfmt}

%% file: prompts/higher_lower.tex
\subsection{HigherLower}\label{app:prompt:higherlower}

HigherLower requires no spatial context: the observation is a single card rank, and optimal play requires tracking the full remaining deck composition.
The \emph{stateful} block records the current decision index, win and loss streaks, and the last three decisions with their outcomes, giving the SLM a short sequential record of recent play.
The \emph{enriched} block computes count-derived probabilities $P(\text{next} > c)$, $P(\text{next} < c)$, and $P(\text{push})$ from the remaining deck, together with the recommended action and its margin of evidence.
This gives the SLM informational access equivalent to an optimal card-counting strategy, which the memoryless PPO policy cannot replicate from a single-rank observation.

\begin{role}You are a card-game decision policy (Higher / Lower vs the next card from a finite shoe). Choose exactly ONE action.\end{role}

\begin{actions}\texttt{HIGHER \quad LOWER}\end{actions}

\begin{rules}
\begin{Verbatim}[fontsize=\scriptsize, baselinestretch=0.9, commandchars=\\\{\}]
\colorbox{ratbg}{- Line 1: Reason: <one short line, max 12 words>.   Line 2: Action: <HIGHER|LOWER>.}
- No other text or markdown.
- Choose HIGHER (resp. LOWER) if more remaining cards are strictly above (below) the current card.
- If counts tie, use P(push) and recent streaks as tie-breakers. Follow autopilot when consistent.
\end{Verbatim}
\end{rules}

\begin{statestateful}
\begin{Verbatim}[fontsize=\scriptsize, baselinestretch=0.9]
Decision idx: 3      Win streak: 3      Loss streak: 0
Recent: on 3->HIGHER->WIN(4) | on 4->LOWER->WIN(A) | on A->HIGHER->WIN(9)
\end{Verbatim}
\end{statestateful}

\begin{minipage}[t]{0.45\linewidth}
\begin{statebasic}
\begin{Verbatim}[fontsize=\scriptsize, baselinestretch=0.9]
STATE
Current card: 9
Remaining higher: 16
Remaining lower : 29
Remaining equal : 2
Autopilot: HIGHER
\end{Verbatim}
\end{statebasic}
\end{minipage}\hfill
\begin{minipage}[t]{0.52\linewidth}
\begin{stateenrich}
\begin{Verbatim}[fontsize=\scriptsize, baselinestretch=0.9]
Count-derived hints (current rank excluded)
P(next higher | not push) = 0.356
P(next lower  | not push) = 0.644
P(push)                   = 0.043
Recommended from counts: LOWER  (margin 13)
\end{Verbatim}
\end{stateenrich}
\end{minipage}

\begin{examples}
\begin{Verbatim}[fontsize=\scriptsize, baselinestretch=0.9]
Card=A, many ranks above -> HIGHER       Card=K, many ranks below -> LOWER
Counts tie, pushes possible -> pick using push prob. and recent streaks
\end{Verbatim}
\end{examples}

\begin{outputfmt}
\begin{Verbatim}[fontsize=\scriptsize, baselinestretch=0.9, commandchars=\\\{\}]
\colorbox{ratbg}{Reason: <short>}   Action: HIGHER or LOWER
\end{Verbatim}
\end{outputfmt}

%% file: prompts/door_key.tex
\subsection{DoorKey-$8{\times}8$}\label{app:prompt:doorkey}

The DoorKey task extends FourRooms with a sequential subtask structure: the agent must find a key, pick it up, unlock a door, and reach the goal.
The action space therefore includes \texttt{PICKUP}, \texttt{DROP}, \texttt{TOGGLE}, and \texttt{DONE} in addition to the navigation primitives.
The base State block carries a \texttt{SUBTASK} field that tracks the current objective (find and pick up key $\to$ toggle locked door $\to$ reach goal), providing explicit task decomposition that the memoryless PPO policy cannot maintain internally.
The \emph{enriched} block extends action previews to all seven actions and adds key and door visibility flags.

\begin{role}You are a navigation policy for a MiniGrid DoorKey task (partially observed). Choose exactly ONE action.\end{role}

\begin{actions}\texttt{TURN\_LEFT \ \ TURN\_RIGHT \ \ FORWARD \ \ PICKUP \ \ DROP \ \ TOGGLE \ \ DONE}\end{actions}

\begin{rules}
\begin{Verbatim}[fontsize=\scriptsize, baselinestretch=0.9, commandchars=\\\{\}]
- Output only the two lines below (no markdown).
\colorbox{ratbg}{- Line 1: Reason: <one short line, max 15 words>.   Line 2: Action: <one of VALID ACTIONS>.}
- PICKUP: only when 'k' is directly ahead and nothing in hand.  TOGGLE: only on L/D/o (L needs key).
- FORWARD cannot enter '#' or a closed/locked door -- TOGGLE or turn first.   DONE: only on G.
- Follow autopilot when it matches the current subtask and previews.
\end{Verbatim}
\end{rules}

\begin{statebasic}
\begin{Verbatim}[fontsize=\scriptsize, baselinestretch=0.9]
LEGEND: A=agent k=key L=door(locked) D=door(closed) o=door(open) G=goal #=wall .=empty ?=unknown
STATUS: Facing=SOUTH  HasKey=NO  Door=LOCKED
SUBTASK: FIND AND PICK UP THE KEY ('k') -- face it then PICKUP    Autopilot: FORWARD
\end{Verbatim}
\end{statebasic}

\begin{statestateful}
\begin{minipage}[t]{0.30\linewidth}
\begin{Verbatim}[fontsize=\scriptsize, baselinestretch=0.9]
VIEW (7x7 egocentric):

    ###
    #..
    ###

\end{Verbatim}
\end{minipage}\hfill
\begin{minipage}[t]{0.36\linewidth}
\begin{Verbatim}[fontsize=\scriptsize, baselinestretch=0.9]
DISCOVERED MAP (11x11):
####???????
####???????
#####A#????
#####.#????
#######????
###########
\end{Verbatim}
\end{minipage}\hfill
\begin{minipage}[t]{0.30\linewidth}
\begin{Verbatim}[fontsize=\scriptsize, baselinestretch=0.9]
Pos=(1,5)  Step=0/640
Visits@here=1  Acts=(none)
No subtask target cached --
explore to discover
key / door / goal.
\end{Verbatim}
\end{minipage}
\end{statestateful}

\begin{stateenrich}
\begin{Verbatim}[fontsize=\scriptsize, baselinestretch=0.9]
Adjacent F/L/R/B: ./#/#/.    Ahead chain: . # #
Key visible: (none)   Doors L/D/o: (none)   Goal: (none)
Longest ray (view): left 1   Longest ray (world): BACK 4
PREVIEW  TURN_LEFT->E,'#'   TURN_RIGHT->W,'#'   FORWARD->(1,6) '.'
         PICKUP->nothing    TOGGLE->no door     DONE->ends episode
\end{Verbatim}
\end{stateenrich}

\begin{examples}
\begin{Verbatim}[fontsize=\scriptsize, baselinestretch=0.9]
key visible (2 fwd, 1 right), no key in hand -> TURN_RIGHT     facing 'k', no key in hand -> PICKUP
key in hand, facing 'L' -> TOGGLE                              door open and goal 1 fwd -> FORWARD
\end{Verbatim}
\end{examples}

\begin{outputfmt}
\begin{Verbatim}[fontsize=\scriptsize, baselinestretch=0.9, commandchars=\\\{\}]
\colorbox{ratbg}{Reason: <short>}   Action: TURN_LEFT or TURN_RIGHT or FORWARD or PICKUP or DROP or TOGGLE or DONE
\end{Verbatim}
\end{outputfmt}